\documentclass[conference,letterpaper]{IEEEtran}
\usepackage[T1]{fontenc}
\usepackage[utf8]{inputenc}
\usepackage{cite}
\usepackage{amsmath,amssymb}
\usepackage{graphicx}
\usepackage{tikz}
\usetikzlibrary{arrows.meta,shapes.geometric}
\usepackage{booktabs,tabularx,array}
\usepackage{url}
\usepackage[hidelinks]{hyperref}
\hypersetup{pdftitle={Same-Number Citation Swaps: Stress-Testing Jev as a Financial Evidence Judge},pdfauthor={Chuhong Xu, Bo Su, Ziyao Chen, Ruiyang Xu, Shimeng Dai, Xinyu Qiu},pdfsubject={Agenthon CR8 scientific content; six individual author blocks; October 5, 2026}}
\begin{document}
\title{Same-Number Citation Swaps: Stress-Testing Jev as a Financial Evidence Judge}
\author{
\IEEEauthorblockN{\makebox[0.30\textwidth]{\href{https://orcid.org/0009-0007-3447-2558}{Chuhong Xu}}}
\IEEEauthorblockA{Sofia University\\
chuhong.xu@sofia.edu}
\and
\IEEEauthorblockN{\makebox[0.30\textwidth]{\href{https://orcid.org/0009-0009-8023-8134}{Bo Su}}}
\IEEEauthorblockA{Indiana University\\
subo@iu.edu}
\and
\IEEEauthorblockN{\makebox[0.30\textwidth]{\href{https://orcid.org/0009-0003-8577-1487}{Ziyao Chen}}}
\IEEEauthorblockA{University of California, San Diego\\
cziyao@ucsd.edu}\and[\hfill\mbox{}\par\vspace{\baselineskip}\mbox{}\hfill]
\IEEEauthorblockN{\makebox[0.30\textwidth]{\href{https://orcid.org/0000-0002-4973-8458}{Ruiyang Xu}}}
\IEEEauthorblockA{Northeastern University\\
xu.r@northeastern.edu}
\and
\IEEEauthorblockN{\makebox[0.30\textwidth]{\href{https://orcid.org/0009-0009-9260-0401}{Shimeng Dai}}}
\IEEEauthorblockA{Michigan State University\\
daishime@msu.edu}
\and
\IEEEauthorblockN{\makebox[0.30\textwidth]{\href{https://orcid.org/0009-0005-3263-8505}{Xinyu Qiu}}}
\IEEEauthorblockA{Northeastern University\\
qiu.xiny@northeastern.edu}
}
\IEEEoverridecommandlockouts
\makeatletter
\IEEEaftertitletext{\ifx\@author\@empty\vspace{-2\baselineskip}\fi}
\makeatother
\maketitle
\begin{abstract}
Financial reports repeat values across periods, metrics and accounting lines, allowing an LLM-generated calculation to be numerically correct while citing the wrong financial role. We evaluate what probabilistic evidence verification adds beyond number matching using Jev as a source-support verifier for GPT-4.1-mini calculation traces. A signed-number-at-pointer baseline explains most recovery over exact quotation checks. To isolate the remaining role-recognition problem, we hold operands and arithmetic fixed, move citations between same-number cells, and retain controls that express equivalent facts. These contrasts reveal both wrong-role citations that pass and valid alternative citations that are withheld. Explicit column labels improve selected wrong-role decisions while also lowering support for some equivalent evidence. A constructed follow-up on 36 new source pages, labeled by a non-author reviewer, extends this evaluation and exposes the same tradeoff between detecting role errors and retaining valid citations. The contribution is a controlled evaluation that identifies what a probabilistic financial verifier distinguishes when numerical matching is held fixed. For LLM-based financial assistants, it makes numerical correctness, cited-role support and acceptance outcomes separately assessable.
\end{abstract}
\vspace{\baselineskip}
\section{Introduction}\label{sec:1}

An LLM-based financial research assistant produces a numerical calculation and citations for its operands. Verifying the output requires checking both the calculation and the cited cells' period, metric and accounting role. In our GPT-4.1-mini records, 27 of 192 traces have executed outputs that match the answer key yet cite unsupported roles. A literal quotation check already catches all 27, making it essential to compare probabilistic verification against deterministic citation checks. We ask what the verifier adds once numerical citation matching is in place, and whether it retains valid evidence while detecting role errors.

FinQA supplies numerical reasoning questions over financial reports \cite{r1}. Its answers come with gold programs, but a number and its cited basis still have to be assessed separately. FinanceBench pairs financial answers with supporting evidence \cite{r2}. Financial reports repeat the same value across years, metrics and accounting categories, so a value found at a citation may belong to the wrong role, while a different citation may express the same fact. We ask: what does probabilistic financial evidence verification add beyond numerical matching, and which citation changes reveal its limits?

We use Jev to verify GPT-4.1-mini calculation traces, each containing operands, an operation, source pointers and an answer. We resolve each pointer into the supplied excerpt and ask Jev whether the cited evidence supports the calculation. This source-support probability concerns a different event from numerical answer selection.

The experiments build on one another. Natural traces give performance against retained source labels, and a signed-number-at-pointer rule measures the contribution of number matching. Same-number swaps then hold numerical agreement fixed to test financial-role recognition, with equivalent-fact controls distinguishing harmless citation changes from genuine role errors. Explicit column rendering probes selected failures while preserving the source. A follow-up repeats the design on new pages selected using source information alone, with non-author source review.

The evaluation measures incremental value as supported acceptances and joint passes relative to signed-number matching, financial-role discrimination at fixed numerical agreement, and representation sensitivity under explicit column rendering. Equivalent-fact controls measure retention of valid evidence. The 192 natural records are LLM outputs; the 36-question follow-up uses program-derived records. Historical assistant labels and returned reviewer labels remain distinct throughout. Together, these comparisons characterize a probabilistic verification component for financial assistants.

Section~\ref{sec:4} reports the original, prospectively specified release endpoint. Component corrections did not improve fixed-gate released answers on this bank; the later source-support evaluation addresses the evidence-verification component.

\section{Financial records and measurement targets}\label{sec:2}

\subsection{Source population and calculation traces}\label{sec:2.1}

The bank contains 96 FinQA test questions with single-operation calculations. The source is fixed to FinQA revision 0f16e286 \cite{r1}. Of 1,147 test records, 243 pass the structural screen, spanning 175 source pages. Source-only assistant review of the first 144 hash-ranked pages accepts 115. We additionally exclude a known cross-year duplicate of a reserved repair-pilot question, then take the first 96 remaining accepted questions. They cover 96 pages and 88 company/year reports. Retained screening decisions and reasons define the study population: selected single-operation questions from the benchmark.

Before collection, we committed the sample, prompts, execution rules, original release policies and analysis specification. Each question receives one Jev numerical selection and two GPT-4.1-mini traces from identical menu-free inputs. Jev selects among five numerical candidates: the rounded key, one numerical neighbor and three arithmetic distractors, assigned to neutral IDs. A fixed rule consults a trace only when the maximum Choice probability is strictly below 0.8; this selector role defines the bank's original release endpoint (Section~\ref{sec:4}), while the later verifier role studied in Section~\ref{sec:3} evaluates cited evidence. All 288 requests succeed on their first attempt. Both draws are retained, yielding the 192 natural records. Later source-probability and citation experiments were designed after inspecting these records: retrospective extensions of a prospectively collected bank.

External calculation is established in PAL \cite{r5}. Checking an operand's financial role requires linking that operand to its source context. Financial QA agents also use external calculation \cite{r6}.

Each trace requests two decimal operands, their source pointers and quotations, one arithmetic operation, a stated answer, an output unit and a justification. A bounded executor applies predefined addition, subtraction, multiplication or division to the supplied operands at their emitted scale, then rounds to the frozen question precision. Missing-operand inference, value rescaling and generated-code execution are outside its scope. Stated and executed numbers are paired readings of the same trace. The model receives the question and supplied financial excerpt; the numerical key, gold reasoning program and source annotations are withheld.

\subsection{Numerical correctness, evidence support and acceptance}\label{sec:2.2}

Numerical correctness means matching the frozen rounded FinQA-derived key. Mismatches and unavailable outputs are separate outcomes. We assess numerical correctness and source support as separate dimensions: source support concerns the cited operands' financial roles, while numerical correctness also depends on arithmetic and output scaling.

Cited-role support asks whether the actual cited cells, rows, headers or text spans provide the operands in the roles required by the question: metric, entity or cohort, period, status and contextual sign. It also includes task-role completeness; for example, a cash-spent calculation requires evidence beyond share counts. Arithmetic order and final unit conversion are assessed separately from the cited roles. Assistant review assigns supported, unsupported, ambiguous or unassessable labels and retains a source-linked reason. An additional full-excerpt judgment asks whether the expressed calculation is recoverable elsewhere in the supplied evidence. Cited-role support remains anchored to the actual emitted pointer.

The natural-record source labels come from unblinded assistant review of all 192 traces, with documented adjudications and two unresolved records retained. Support is defined by the cited source context; numerical outcomes and later verifier probabilities are separate readouts. Independent human confirmation of these natural-record labels remains incomplete. The returned non-author review in Section~\ref{sec:3.4} applies exclusively to the new constructed contrasts.

Literal correspondence requires the emitted quotation and numerical operand span to agree with the location addressed by the pointer. Combined with available execution and matching output units, this forms the literal/unit acceptance rule. Acceptance is a decision under a stated rule, distinct from source support and from numerical correctness. At a high acceptance cutoff, a withheld citation can still carry a support probability above 0.5. A joint pass requires both supported cited roles and an executed key match.

Evidence binding refers here to resolving a citation to its actual source location and presenting that location's context. The initial verifier input includes the complete source table, visible row/column indices, the table's header row and the cited row. Thus column identity is already recoverable. The later intervention repeats the selected column's header explicitly alongside the citation, varying its representation while holding source availability fixed.

\subsection{Models and reporting units}\label{sec:2.3}

All Jev calls return {\small\nolinkurl{typesafe/jev-1.13-20260917}} through OpenRouter with the TypeSafe provider. Numerical answer selection uses Choice; source verification uses one independently specified binary Noul event. Choice and Noul probabilities are analyzed as distinct measurement targets. Analyses use the exact returned probabilities, preserving their original scale; the interface's native confidence statistic is excluded from accuracy estimation.

GPT-4.1-mini uses the OpenAI provider, disabled provider fallback, temperature 0, top-p 1 and a 1,024-token output limit for traces. The prompt requests JSON, and decoding is unconstrained. The returned {\small\nolinkurl{openai/gpt-4.1-mini}} identifier is a serving alias; it does not pin immutable model weights. Jev selection and menu-free trace generation have different tasks and output budgets, defining a workflow comparison between their assigned roles.

Natural-record summaries retain all traces and question identities. Verifier and original-release resampling runs use 2,000 and 20,000 draws, respectively. Question bootstrap intervals quantify variation across sampled questions while retaining both draws. Annotation uncertainty and future serving variation lie outside their coverage. Constructed citation studies report contrasts and source questions separately; repeated identical inputs are counted as technical calls within the same semantic case.

\section{Verification experiments}\label{sec:3}

\subsection{Natural records: discrimination and a stronger numerical control}\label{sec:3.1}

The natural records contain both numerical and citation errors (Table~\ref{tab:1}). Of 192 executed outputs, 170 match the key, 21 do not, and one is unavailable. Source review finds 153 supported traces, 37 unsupported, one ambiguous and one unassessable. Twenty-seven key-matching traces on 14 questions cite unsupported roles even though their derivations are recoverable elsewhere in the excerpt. Literal correspondence detects all 27, demonstrating why numerical correctness and cited-role support require separate assessment. Conversely, literal correspondence rejects 33 supported traces. The combined literal/unit rule accepts two unsupported-role traces; both are numerical mismatches.

\begin{table*}[!t]
\centering
\caption{Numerical outputs and cited-role diagnostics on the fresh bank.}
\label{tab:1}
\small
\setlength{\tabcolsep}{5pt}
\renewcommand{\arraystretch}{1.12}
\par\smallskip\noindent \textbf{A. Numerical outputs.} The 96 questions are retained in each draw. These are all-trace component results, not the selected-gate release outcomes in Table~\ref{tab:3}.\par\smallskip
\begin{tabularx}{\textwidth}{@{}>{\raggedright\arraybackslash}X>{\raggedright\arraybackslash}X>{\raggedright\arraybackslash}X@{}}
\toprule
\textbf{Trace draw} & \textbf{Output} & \textbf{Matches / mismatches / unavailable} \\
\midrule
0 & Stated number & 79 / 16 / 1 \\
0 & Executed number & 84 / 11 / 1 \\
0 & Execution + lexical/unit acceptance & 58 / 3 / 35 \\
1 & Stated number & 80 / 16 / 0 \\
1 & Executed number & 86 / 10 / 0 \\
1 & Execution + lexical/unit acceptance & 59 / 2 / 35 \\
\bottomrule
\end{tabularx}

\par\smallskip\noindent \textbf{B. Cited-role and literal-check diagnostics.} All draw cells retain the scheduled denominator. The question column counts at least one qualifying draw; its rows need not be mutually exclusive. Labels are assistant judgments.\par\smallskip
\begin{tabularx}{\textwidth}{@{}>{\raggedright\arraybackslash}Xrrrr@{}}
\toprule
\textbf{Measurement} & \textbf{Draw 0} & \textbf{Draw 1} & \textbf{All draws} & \textbf{Questions} \\
\midrule
Cited roles supported & 74/96 & 79/96 & 153/192 & 79/96 \\
Cited roles unsupported & 20/96 & 17/96 & 37/192 & 20/96 \\
Cited roles ambiguous & 1/96 & 0/96 & 1/192 & 1/96 \\
Cited roles unassessable & 1/96 & 0/96 & 1/192 & 1/96 \\
Key-matching executed number, unsupported cited roles & 14/96 & 13/96 & 27/192 & 14/96 \\
Supported cited roles, literal correspondence fails & 15/96 & 18/96 & 33/192 & 19/96 \\
Literal/unit accepted, unsupported cited roles & 2/96 & 0/96 & 2/192 & 2/96 \\
\bottomrule
\end{tabularx}

\end{table*}

We next ask whether a probabilistic source verifier can improve this acceptance tradeoff. Every natural-record, citation-swap and rendering request uses the same fixed Noul instruction, beginning: "Does the cited evidence support the proposed record's operand values and financial roles for the quantity requested in the question?" The instruction requires support at the actual cited locations for the requested metric or entity, period, sign and source unit, allowing harmless paraphrases and using surrounding context. It explicitly excludes rejection based solely on a wrong arithmetic operator, operand order, final rounding or missing final display-unit conversion. The same instruction is present in all 590 saved evaluation requests, including the fresh-source follow-up in Section~\ref{sec:3.4}.

Only the allowlisted source and record state varies. Jev receives the full supplied excerpt, question, requested unit and precision, operation, operands, justification, executed value and deterministically resolved citations. The numerical key, source labels, checker verdicts and original stated-answer field are absent. This input defines the source-support target while leaving numerical-correctness cues available to the model. We wrote the predicate after inspecting the retained labels and tested the input format in four development checks; all 192 natural-record requests then succeeded. The experiment evaluates a new probability target retrospectively on known records.

On the 190 binary labels, the source probabilities have Brier score 0.03378 and AUROC 0.98587. Question-bootstrap 95\% intervals from 2,000 resamples are [0.01814, 0.05260] and [0.96356, 0.99938], respectively. The binary support rate of 0.8053 is close to the mean probability of 0.7975. Calibration also requires agreement within probability ranges: in the largest fixed probability bin, 146 records have mean probability 0.9375 and observed support rate 0.9932. A pre-inference sensitivity excluding six task-role boundary records from questions 467, 576 and 991 retains 184 binary records, with Brier 0.03132 and AUROC 0.98551. Original labels remain unchanged.

At the inclusive 0.8 source-probability cutoff, all 27 key-matching traces with unsupported cited roles are withheld (maximum probability 0.79), as are nine of the ten unsupported traces with numerical mismatches. Thus no correct-number/unsupported-role trace is accepted at that cutoff in this natural bank. Section~\ref{sec:3.2} tests role recognition with numerical agreement held fixed.

To measure the increment beyond number matching, we add a post hoc zero-call control, the relaxed pointer rule: remove exact-quotation matching while still requiring each signed operand value at its actual cited location. Matching is confined to the string at the emitted table-cell or text-span pointer, using the operand's original sign and scale. Whole-row list pointers are ineligible. Numerical spans are matched as complete decimal values, ignoring thousands separators while preserving accounting parentheses as a negative sign. The shared release guards require an available execution and exact equality of the emitted and requested unit strings. This guard checks unit labels; scaling correctness is a separate numerical question. Both rules operate on the source and emitted record, with answer keys and support labels excluded.

Table~\ref{tab:2} retains all four inclusive cutoffs listed before verifier inference: 0.5, 0.8, 0.9 and 0.95. New releases require available execution and matching output units in addition to the probability gate. The OR policy rescues records rejected by the original literal/unit rule. Unresolved labels remain in the full denominator rather than becoming supported records.

\begin{table*}[!t]
\centering
\caption{Source-probability gates on the existing 192 traces. All thresholds are inclusive. Supported/unsupported/unresolved refer to retained assistant source-role labels; joint passes additionally match the numerical key.}
\label{tab:2}
\small
\setlength{\tabcolsep}{5pt}
\renewcommand{\arraystretch}{1.12}
\begin{tabularx}{\textwidth}{@{}>{\raggedright\arraybackslash}Xrrrrr@{}}
\toprule
\textbf{Acceptance rule} & \textbf{Accepted / 192} & \textbf{Supported} & \textbf{Unsupported} & \textbf{Unresolved} & \textbf{Joint passes} \\
\midrule
Literal/unit rule & 122 & 120 & 2 & 0 & 117 \\
Relaxed pointer rule (post hoc) & 154 & 151 & 2 & 1 & 141 \\
Jev source gate, 0.5 & 159 & 152 & 6 & 1 & 143 \\
Jev source gate, 0.8 & 146 & 145 & 1 & 0 & 142 \\
Jev source gate, 0.9 & 126 & 125 & 1 & 0 & 123 \\
Jev source gate, 0.95 & 87 & 87 & 0 & 0 & 87 \\
Literal OR Jev, 0.5 & 161 & 153 & 7 & 1 & 143 \\
Literal OR Jev, 0.8 & 151 & 149 & 2 & 0 & 143 \\
Literal OR Jev, 0.9 & 144 & 142 & 2 & 0 & 137 \\
Literal OR Jev, 0.95 & 135 & 133 & 2 & 0 & 130 \\
\bottomrule
\end{tabularx}

\end{table*}

This relaxed pointer rule recovers 31 of the 33 supported literal failures and 24 of the 26 that also match the key, with no additional unsupported acceptance and one ambiguous acceptance.

At 0.8, Jev accepts 146 records with 142 joint passes, compared with 154 accepted and 141 joint passes for the relaxed rule. The tradeoff is fewer accepted records for one additional joint pass. The two Jev-only joint passes both belong to question 218, where a negative benefit is used as a positive magnitude in a subtraction. This contextual-sign case identifies the specific distinction from strict signed-number matching. As a hard 0/1 classifier, the relaxed rule also has lower Brier score on these labels (4/190 = 0.02105), although its erroneous certainties have infinite unregularized log loss. These comparisons motivate testing role recognition while holding numerical matches fixed.

Paired decision differences. At the same inclusive 0.8 cutoff and with identical release guards, both rules accept 144 traces, Jev alone accepts two, the numerical rule alone accepts ten, and both withhold 36. Their joint-pass counts in these groups are 140, two, one and zero, respectively. The net gain of one joint pass is therefore two recoveries on question 218 minus one withheld valid trace on question 583, draw 1. Removing both question-218 traces leaves 190 records: Jev has 140 joint passes versus 141 for the numerical rule. This leave-one-question-out sensitivity locates the observed increment in a single contextual-sign case.

Seven of the ten rule-only acceptances have supported cited roles but mismatched numerical answers, spanning four questions. They explain the difference between ten such accepted traces under the rule and three under Jev. Withholding these records can improve accepted-answer precision while leaving the joint-pass count unchanged. The pattern links lower source probabilities to broader calculation defects. Because arithmetic, explanations and other input features vary together in these natural traces, their individual contributions remain unresolved.

\subsection{Same-number citations: isolating financial role}\label{sec:3.2}

Systematic tabular-reasoning probes use evidence perturbations to examine model behavior \cite{r11}. We apply this logic to a source-support verifier while holding operands and arithmetic fixed. TabDSR also uses table-data perturbations \cite{r12}. The present intervention changes the citation pointer and then the presentation of its selected column.

In each matched contrast, arm A retains the original citation and arm B points one operand to a different cell containing the same numerical value. Both arms use exact cell quotations and a neutral justification. The question, complete source, operand values, operation and executed result remain fixed. The changed pointer and its resolved context are the intended difference. Both arms therefore satisfy the same signed-number-at-pointer criterion.

Source review before inference separates clear wrong-role changes from equivalent facts and unresolved boundaries. Equivalent-fact labels require support from the source context as well as equal values. For example, an ending balance and the next period's opening balance may represent the same carried-forward amount. Original proposed classes and subsequent pre-inference adjudications are both saved. Categories remain fixed throughout inference.

Same value, wrong period (question 906). The question asks for Intel's cumulative return relative to the Dow Jones U.S. Technology Index in 2010. The record divides Intel's 157 by the index's 191. Arm A cites {\small\nolinkurl{/table/2/3}}, headed "2010"; arm B cites {\small\nolinkurl{/table/2/4}}, headed "2011". Both index cells contain "\$ 191", and the executed ratio remains 0.82. Numerical matching is preserved, while the swapped citation supplies the following year's index fact. The citation contract requires the period requested by the question.

Different headings, equivalent fact (question 535). For December 2012 repurchases, arm A cites 102,400 total shares at {\small\nolinkurl{/table/3/1}}; arm B cites the same count under publicly announced plans or programs at {\small\nolinkurl{/table/3/3}}. The preamble says the quarter's repurchases were "pursuant to our publicly announced stock repurchase program", and note (1) attaches the total-purchase column to that program. This source statement, rather than equality alone, supports the equivalent-purchase-set label. Both arms retain the December average-price operand. The judgment concerns cited roles; fee completeness is a separate calculation question. Both examples include the assistant judgments and their source-based rationales for inspection.

\begin{figure*}[!t]
\centering
\begingroup%
\definecolor{crInk}{HTML}{303942}%
\definecolor{crBlue}{HTML}{0072B2}%
\definecolor{crOrange}{HTML}{D55E00}%
\definecolor{crLight}{HTML}{F1F4F6}%
\definecolor{crGrid}{HTML}{DAE0E4}%
\begin{tikzpicture}[x=1pt,y=1pt,every node/.style={font=\fontsize{8}{9}\selectfont,text=crInk,inner sep=0pt,outer sep=0pt}]%
\path[use as bounding box] (0,0) rectangle (504,235);%
\draw[draw=crInk,fill=crInk,line width=.9pt] (177.000,227.000) circle (2.3pt);%
\node[anchor=west,text=crInk,font=\fontsize{8}{9}\selectfont] at (183.000,227.000) {A: original citation};%
\draw[draw=crBlue,fill=white,line width=.9pt] (277.750,224.750) rectangle (282.250,229.250);%
\node[anchor=west,text=crInk,font=\fontsize{8}{9}\selectfont] at (287.000,227.000) {B: same-number swap};%
\draw[draw=crInk,line width=0.8pt,densely dashed] (1.000,227.000) -- (17.000,227.000);%
\node[anchor=west,text=crInk,font=\fontsize{8}{9}\selectfont] at (22.000,227.000) {Accept mean $\geq 0.8$};%
\node[anchor=west,text=crInk,font=\fontsize{8}{9}\selectfont] at (0.000,213.000) {Question / interpretation / changed coordinate};%
\node[anchor=center,text=crInk,font=\fontsize{8}{9}\selectfont] at (423.000,213.000) {A mean};%
\node[anchor=center,text=crInk,font=\fontsize{8}{9}\selectfont] at (479.000,213.000) {B mean};%
\fill[crLight] (0.000,132.375) rectangle (504.000,202.875);%
\fill[crLight] (0.000,26.625) rectangle (504.000,61.875);%
\draw[draw=crGrid,line width=0.45pt] (178.000,27.000) -- (178.000,205.000);%
\draw[draw=crInk,line width=0.5pt] (178.000,27.000) -- (178.000,25.000);%
\node[anchor=center,text=crInk,font=\fontsize{8}{9}\selectfont] at (178.000,18.000) {0};%
\draw[draw=crGrid,line width=0.45pt] (220.000,27.000) -- (220.000,205.000);%
\draw[draw=crInk,line width=0.5pt] (220.000,27.000) -- (220.000,25.000);%
\node[anchor=center,text=crInk,font=\fontsize{8}{9}\selectfont] at (220.000,18.000) {.2};%
\draw[draw=crGrid,line width=0.45pt] (262.000,27.000) -- (262.000,205.000);%
\draw[draw=crInk,line width=0.5pt] (262.000,27.000) -- (262.000,25.000);%
\node[anchor=center,text=crInk,font=\fontsize{8}{9}\selectfont] at (262.000,18.000) {.4};%
\draw[draw=crGrid,line width=0.45pt] (304.000,27.000) -- (304.000,205.000);%
\draw[draw=crInk,line width=0.5pt] (304.000,27.000) -- (304.000,25.000);%
\node[anchor=center,text=crInk,font=\fontsize{8}{9}\selectfont] at (304.000,18.000) {.6};%
\draw[draw=crGrid,line width=0.45pt] (346.000,27.000) -- (346.000,205.000);%
\draw[draw=crInk,line width=0.5pt] (346.000,27.000) -- (346.000,25.000);%
\node[anchor=center,text=crInk,font=\fontsize{8}{9}\selectfont] at (346.000,18.000) {.8};%
\draw[draw=crGrid,line width=0.45pt] (388.000,27.000) -- (388.000,205.000);%
\draw[draw=crInk,line width=0.5pt] (388.000,27.000) -- (388.000,25.000);%
\node[anchor=center,text=crInk,font=\fontsize{8}{9}\selectfont] at (388.000,18.000) {1};%
\draw[draw=crInk,line width=0.5pt] (178.000,27.000) -- (388.000,27.000);%
\node[anchor=center,text=crInk,font=\fontsize{8}{9}\selectfont] at (283.000,6.000) {Mean source-support probability};%
\draw[draw=crInk,line width=0.8pt,densely dashed] (346.000,27.000) -- (346.000,205.000);%
\node[anchor=west,text=crInk,font=\fontsize{8}{9}\selectfont] at (1.000,197.000) {132 / wrong period / row};%
\draw[draw=crInk,line width=0.7pt] (382.750,197.000) -- (201.100,197.000);%
\draw[draw=crInk,fill=crInk,line width=.9pt] (382.750,198.100) circle (2.3pt);%
\draw[draw=crBlue,fill=white,line width=.9pt] (198.850,193.650) rectangle (203.350,198.150);%
\node[anchor=center,text=crInk,font=\fontsize{8}{9}\selectfont] at (423.000,197.000) {0.975};%
\node[anchor=center,text=crBlue,font=\fontsize{8}{9}\selectfont] at (479.000,197.000) {0.110};%
\node[anchor=west,text=crInk,font=\fontsize{8}{9}\selectfont] at (1.000,185.250) {906 / wrong period / column};%
\draw[draw=crInk,line width=0.7pt] (378.550,185.250) -- (377.500,185.250);%
\draw[draw=crInk,fill=crInk,line width=.9pt] (378.550,186.350) circle (2.3pt);%
\draw[draw=crBlue,fill=white,line width=.9pt] (375.250,181.900) rectangle (379.750,186.400);%
\node[anchor=center,text=crInk,font=\fontsize{8}{9}\selectfont] at (423.000,185.250) {0.955};%
\node[anchor=center,text=crBlue,font=\fontsize{8}{9}\selectfont] at (479.000,185.250) {0.950};%
\node[anchor=west,text=crInk,font=\fontsize{8}{9}\selectfont] at (1.000,173.500) {930/period / wrong period / column};%
\draw[draw=crInk,line width=0.7pt] (380.650,173.500) -- (377.500,173.500);%
\draw[draw=crInk,fill=crInk,line width=.9pt] (380.650,174.600) circle (2.3pt);%
\draw[draw=crBlue,fill=white,line width=.9pt] (375.250,170.150) rectangle (379.750,174.650);%
\node[anchor=center,text=crInk,font=\fontsize{8}{9}\selectfont] at (423.000,173.500) {0.965};%
\node[anchor=center,text=crBlue,font=\fontsize{8}{9}\selectfont] at (479.000,173.500) {0.950};%
\node[anchor=west,text=crInk,font=\fontsize{8}{9}\selectfont] at (1.000,161.750) {388 / wrong cohort / row};%
\draw[draw=crInk,line width=0.7pt] (371.200,161.750) -- (259.900,161.750);%
\draw[draw=crInk,fill=crInk,line width=.9pt] (371.200,162.850) circle (2.3pt);%
\draw[draw=crBlue,fill=white,line width=.9pt] (257.650,158.400) rectangle (262.150,162.900);%
\node[anchor=center,text=crInk,font=\fontsize{8}{9}\selectfont] at (423.000,161.750) {0.920};%
\node[anchor=center,text=crBlue,font=\fontsize{8}{9}\selectfont] at (479.000,161.750) {0.390};%
\node[anchor=west,text=crInk,font=\fontsize{8}{9}\selectfont] at (1.000,150.000) {746 / wrong period / column};%
\draw[draw=crInk,line width=0.7pt] (369.100,150.000) -- (359.650,150.000);%
\draw[draw=crInk,fill=crInk,line width=.9pt] (369.100,151.100) circle (2.3pt);%
\draw[draw=crBlue,fill=white,line width=.9pt] (357.400,146.650) rectangle (361.900,151.150);%
\node[anchor=center,text=crInk,font=\fontsize{8}{9}\selectfont] at (423.000,150.000) {0.910};%
\node[anchor=center,text=crBlue,font=\fontsize{8}{9}\selectfont] at (479.000,150.000) {0.865};%
\node[anchor=west,text=crInk,font=\fontsize{8}{9}\selectfont] at (1.000,138.250) {930/status / wrong status / column};%
\draw[draw=crInk,line width=0.7pt] (379.600,138.250) -- (377.500,138.250);%
\draw[draw=crInk,fill=crInk,line width=.9pt] (379.600,139.350) circle (2.3pt);%
\draw[draw=crBlue,fill=white,line width=.9pt] (375.250,134.900) rectangle (379.750,139.400);%
\node[anchor=center,text=crInk,font=\fontsize{8}{9}\selectfont] at (423.000,138.250) {0.960};%
\node[anchor=center,text=crBlue,font=\fontsize{8}{9}\selectfont] at (479.000,138.250) {0.950};%
\node[anchor=west,text=crInk,font=\fontsize{8}{9}\selectfont] at (1.000,126.500) {251 / equivalent fact / row};%
\draw[draw=crInk,line width=0.7pt] (379.600,126.500) -- (346.000,126.500);%
\draw[draw=crInk,fill=crInk,line width=.9pt] (379.600,127.600) circle (2.3pt);%
\draw[draw=crBlue,fill=white,line width=.9pt] (343.750,123.150) rectangle (348.250,127.650);%
\node[anchor=center,text=crInk,font=\fontsize{8}{9}\selectfont] at (423.000,126.500) {0.960};%
\node[anchor=center,text=crBlue,font=\fontsize{8}{9}\selectfont] at (479.000,126.500) {0.800};%
\node[anchor=west,text=crInk,font=\fontsize{8}{9}\selectfont] at (1.000,114.750) {535 / equivalent fact / column};%
\draw[draw=crInk,line width=0.7pt] (355.450,114.750) -- (354.400,114.750);%
\draw[draw=crInk,fill=crInk,line width=.9pt] (355.450,115.850) circle (2.3pt);%
\draw[draw=crBlue,fill=white,line width=.9pt] (352.150,111.400) rectangle (356.650,115.900);%
\node[anchor=center,text=crInk,font=\fontsize{8}{9}\selectfont] at (423.000,114.750) {0.845};%
\node[anchor=center,text=crBlue,font=\fontsize{8}{9}\selectfont] at (479.000,114.750) {0.840};%
\node[anchor=west,text=crInk,font=\fontsize{8}{9}\selectfont] at (1.000,103.000) {583 / equivalent fact / column};%
\draw[draw=crInk,line width=0.7pt] (346.000,103.000) -- (346.000,103.000);%
\draw[draw=crInk,fill=crInk,line width=.9pt] (346.000,104.100) circle (2.3pt);%
\draw[draw=crBlue,fill=white,line width=.9pt] (343.750,99.650) rectangle (348.250,104.150);%
\node[anchor=center,text=crInk,font=\fontsize{8}{9}\selectfont] at (423.000,103.000) {0.800};%
\node[anchor=center,text=crBlue,font=\fontsize{8}{9}\selectfont] at (479.000,103.000) {0.800};%
\node[anchor=west,text=crInk,font=\fontsize{8}{9}\selectfont] at (1.000,91.250) {549 / equivalent fact / both};%
\draw[draw=crInk,line width=0.7pt] (377.500,91.250) -- (350.200,91.250);%
\draw[draw=crInk,fill=crInk,line width=.9pt] (377.500,92.350) circle (2.3pt);%
\draw[draw=crBlue,fill=white,line width=.9pt] (347.950,87.900) rectangle (352.450,92.400);%
\node[anchor=center,text=crInk,font=\fontsize{8}{9}\selectfont] at (423.000,91.250) {0.950};%
\node[anchor=center,text=crBlue,font=\fontsize{8}{9}\selectfont] at (479.000,91.250) {0.820};%
\node[anchor=west,text=crInk,font=\fontsize{8}{9}\selectfont] at (1.000,79.500) {9 / equivalent fact / both};%
\draw[draw=crInk,line width=0.7pt] (377.500,79.500) -- (323.950,79.500);%
\draw[draw=crInk,fill=crInk,line width=.9pt] (377.500,80.600) circle (2.3pt);%
\draw[draw=crBlue,fill=white,line width=.9pt] (321.700,76.150) rectangle (326.200,80.650);%
\node[anchor=center,text=crInk,font=\fontsize{8}{9}\selectfont] at (423.000,79.500) {0.950};%
\node[anchor=center,text=crBlue,font=\fontsize{8}{9}\selectfont] at (479.000,79.500) {0.695};%
\node[anchor=west,text=crInk,font=\fontsize{8}{9}\selectfont] at (1.000,67.750) {912 / equivalent fact / both};%
\draw[draw=crInk,line width=0.7pt] (379.600,67.750) -- (247.300,67.750);%
\draw[draw=crInk,fill=crInk,line width=.9pt] (379.600,68.850) circle (2.3pt);%
\draw[draw=crBlue,fill=white,line width=.9pt] (245.050,64.400) rectangle (249.550,68.900);%
\node[anchor=center,text=crInk,font=\fontsize{8}{9}\selectfont] at (423.000,67.750) {0.960};%
\node[anchor=center,text=crBlue,font=\fontsize{8}{9}\selectfont] at (479.000,67.750) {0.330};%
\node[anchor=west,text=crInk,font=\fontsize{8}{9}\selectfont] at (1.000,56.000) {717 / index-baseline boundary / row};%
\draw[draw=crInk,line width=0.7pt] (372.250,56.000) -- (229.450,56.000);%
\draw[draw=crInk,fill=crInk,line width=.9pt] (372.250,57.100) circle (2.3pt);%
\draw[draw=crBlue,fill=white,line width=.9pt] (227.200,52.650) rectangle (231.700,57.150);%
\node[anchor=center,text=crInk,font=\fontsize{8}{9}\selectfont] at (423.000,56.000) {0.925};%
\node[anchor=center,text=crBlue,font=\fontsize{8}{9}\selectfont] at (479.000,56.000) {0.245};%
\node[anchor=west,text=crInk,font=\fontsize{8}{9}\selectfont] at (1.000,44.250) {938 / index-baseline boundary / row};%
\draw[draw=crInk,line width=0.7pt] (379.600,44.250) -- (249.400,44.250);%
\draw[draw=crInk,fill=crInk,line width=.9pt] (379.600,45.350) circle (2.3pt);%
\draw[draw=crBlue,fill=white,line width=.9pt] (247.150,40.900) rectangle (251.650,45.400);%
\node[anchor=center,text=crInk,font=\fontsize{8}{9}\selectfont] at (423.000,44.250) {0.960};%
\node[anchor=center,text=crBlue,font=\fontsize{8}{9}\selectfont] at (479.000,44.250) {0.340};%
\node[anchor=west,text=crInk,font=\fontsize{8}{9}\selectfont] at (1.000,32.500) {785 / accounting/parser boundary / column};%
\draw[draw=crInk,line width=0.7pt] (293.500,32.500) -- (291.400,32.500);%
\draw[draw=crInk,fill=crInk,line width=.9pt] (293.500,33.600) circle (2.3pt);%
\draw[draw=crBlue,fill=white,line width=.9pt] (289.150,29.150) rectangle (293.650,33.650);%
\node[anchor=center,text=crInk,font=\fontsize{8}{9}\selectfont] at (423.000,32.500) {0.550};%
\node[anchor=center,text=crBlue,font=\fontsize{8}{9}\selectfont] at (479.000,32.500) {0.540};%
\end{tikzpicture}%
\endgroup%

\caption{Same-number citation contrasts under the original rendering. A retains the original citation; B cites a different same-number cell. Shading separates wrong roles, equivalent facts and boundaries. Two wrong-role and two equivalent-fact B means fall below the inclusive 0.8 cutoff. Question 930 contributes two contrasts. Means aggregate technical repeats within each arm; counts describe the constructed contrasts.}\label{fig:1}
\end{figure*}
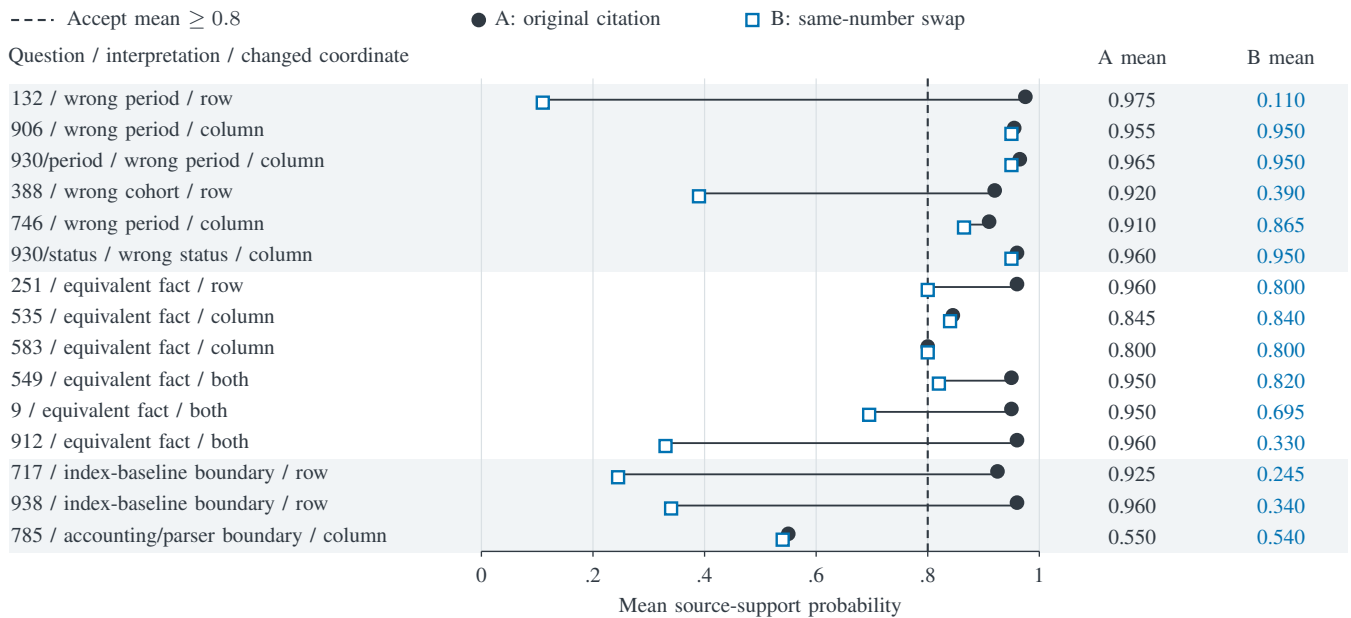

The prepared bank contains 42 evaluation calls over 12 questions; a source-only supplement adds 12 calls over three questions, with question 930 shared. Together they comprise 15 distinct contrasts over 14 questions: six clear wrong-role contrasts over five questions, six equivalent-fact controls and three boundary contrasts. Repeated identical payloads inherited from the natural traces contribute unequal numbers of technical calls. Fig.~\ref{fig:1} averages technical calls within each arm, using the contrast as the semantic reporting unit. All 54 evaluation and eight format-check calls succeed.

Two clear wrong-role B means fall below 0.8; four remain accepted, including wrong-period and wrong-status citations near 0.95. Two of six equivalent-fact B means also fall below 0.8. Boundary cases retain their unresolved status. High aggregate discrimination on natural records therefore coexists with missed role errors and sensitivity to equivalent citations in these constructed cases.

A post hoc structural split suggests a concrete follow-up. Seven contrasts change only the column; their B-minus-A shifts range from -0.045 to zero. Eight change the row, with or without the column; six B means fall below 0.8. Row changes also alter the surrounding numbers and text, so this split motivates a representation test. Column identity is already available through visible pointer indices and the complete table; the next experiment tests the effect of repeating it explicitly in the citation display.

\subsection{Diagnosing selected failures through explicit column labels}\label{sec:3.3}

We reuse all seven column-only contrasts, including four wrong-role contrasts on three questions, two equivalent-fact controls and the accounting/parser boundary. For each A/B arm, the enriched input adds only {\small\nolinkurl{column_header=source.table[0][column_index]}} to every table citation lookup. The header is copied verbatim. Removing the added fields restores the original serialized state exactly. Source content, prompt, model instructions, annotations and numerical values remain fixed.

The inputs and readout are saved before these new calls. Original and enriched renderings are recollected together in one randomized interleaved schedule (seed 80930), with two repetitions of each arm/rendering combination. This schedule supplies contemporaneous original controls for the enriched rendering. All 56 evaluation and four format calls succeed. The seven contrasts cover six source questions and 26 unique evaluation inputs. The two contrasts on question 930 share an A input, producing four technical A calls per rendering for that shared input.

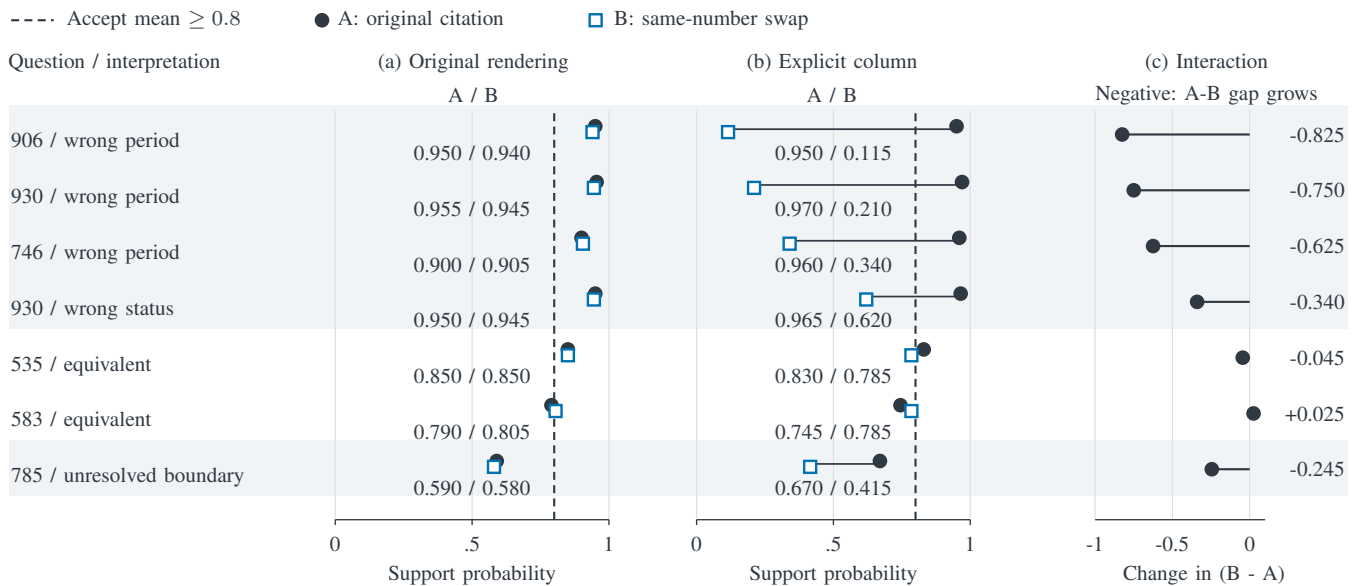
\begin{figure*}[!t]
\centering
\begingroup%
\definecolor{crInk}{HTML}{303942}%
\definecolor{crBlue}{HTML}{0072B2}%
\definecolor{crOrange}{HTML}{D55E00}%
\definecolor{crLight}{HTML}{F1F4F6}%
\definecolor{crGrid}{HTML}{DAE0E4}%
\begin{tikzpicture}[x=1pt,y=1pt,every node/.style={font=\fontsize{8}{9}\selectfont,text=crInk,inner sep=0pt,outer sep=0pt}]%
\path[use as bounding box] (0,0) rectangle (504,223);%
\draw[draw=crInk,fill=crInk,line width=.9pt] (118.000,215.000) circle (2.3pt);%
\node[anchor=west,text=crInk,font=\fontsize{8}{9}\selectfont] at (124.000,215.000) {A: original citation};%
\draw[draw=crBlue,fill=white,line width=.9pt] (218.750,212.750) rectangle (223.250,217.250);%
\node[anchor=west,text=crInk,font=\fontsize{8}{9}\selectfont] at (228.000,215.000) {B: same-number swap};%
\draw[draw=crInk,line width=0.8pt,densely dashed] (1.000,215.000) -- (17.000,215.000);%
\node[anchor=west,text=crInk,font=\fontsize{8}{9}\selectfont] at (22.000,215.000) {Accept mean $\geq 0.8$};%
\node[anchor=west,text=crInk,font=\fontsize{8}{9}\selectfont] at (0.000,199.000) {Question / interpretation};%
\node[anchor=center,text=crInk,font=\fontsize{8}{9}\selectfont] at (175.000,199.000) {(a) Original rendering};%
\node[anchor=center,text=crInk,font=\fontsize{8}{9}\selectfont] at (310.000,199.000) {(b) Explicit column};%
\node[anchor=center,text=crInk,font=\fontsize{8}{9}\selectfont] at (451.000,199.000) {(c) Interaction};%
\node[anchor=center,text=crInk,font=\fontsize{8}{9}\selectfont] at (451.000,187.000) {Negative: A-B gap grows};%
\node[anchor=center,text=crInk,font=\fontsize{8}{9}\selectfont] at (175.000,187.000) {A / B};%
\node[anchor=center,text=crInk,font=\fontsize{8}{9}\selectfont] at (310.000,187.000) {A / B};%
\fill[crLight] (0.000,98.920) rectangle (504.000,182.920);%
\fill[crLight] (0.000,35.920) rectangle (504.000,56.920);%
\draw[draw=crGrid,line width=0.45pt] (123.000,27.000) -- (123.000,181.000);%
\draw[draw=crInk,line width=0.5pt] (123.000,27.000) -- (123.000,25.000);%
\node[anchor=center,text=crInk,font=\fontsize{8}{9}\selectfont] at (123.000,18.000) {0};%
\draw[draw=crGrid,line width=0.45pt] (174.500,27.000) -- (174.500,181.000);%
\draw[draw=crInk,line width=0.5pt] (174.500,27.000) -- (174.500,25.000);%
\node[anchor=center,text=crInk,font=\fontsize{8}{9}\selectfont] at (174.500,18.000) {.5};%
\draw[draw=crGrid,line width=0.45pt] (226.000,27.000) -- (226.000,181.000);%
\draw[draw=crInk,line width=0.5pt] (226.000,27.000) -- (226.000,25.000);%
\node[anchor=center,text=crInk,font=\fontsize{8}{9}\selectfont] at (226.000,18.000) {1};%
\draw[draw=crInk,line width=0.5pt] (123.000,27.000) -- (226.000,27.000);%
\node[anchor=center,text=crInk,font=\fontsize{8}{9}\selectfont] at (174.500,6.000) {Support probability};%
\draw[draw=crInk,line width=0.8pt,densely dashed] (205.400,27.000) -- (205.400,181.000);%
\draw[draw=crGrid,line width=0.45pt] (259.000,27.000) -- (259.000,181.000);%
\draw[draw=crInk,line width=0.5pt] (259.000,27.000) -- (259.000,25.000);%
\node[anchor=center,text=crInk,font=\fontsize{8}{9}\selectfont] at (259.000,18.000) {0};%
\draw[draw=crGrid,line width=0.45pt] (310.500,27.000) -- (310.500,181.000);%
\draw[draw=crInk,line width=0.5pt] (310.500,27.000) -- (310.500,25.000);%
\node[anchor=center,text=crInk,font=\fontsize{8}{9}\selectfont] at (310.500,18.000) {.5};%
\draw[draw=crGrid,line width=0.45pt] (362.000,27.000) -- (362.000,181.000);%
\draw[draw=crInk,line width=0.5pt] (362.000,27.000) -- (362.000,25.000);%
\node[anchor=center,text=crInk,font=\fontsize{8}{9}\selectfont] at (362.000,18.000) {1};%
\draw[draw=crInk,line width=0.5pt] (259.000,27.000) -- (362.000,27.000);%
\node[anchor=center,text=crInk,font=\fontsize{8}{9}\selectfont] at (310.500,6.000) {Support probability};%
\draw[draw=crInk,line width=0.8pt,densely dashed] (341.400,27.000) -- (341.400,181.000);%
\draw[draw=crGrid,line width=0.45pt] (409.000,27.000) -- (409.000,181.000);%
\node[anchor=center,text=crInk,font=\fontsize{8}{9}\selectfont] at (409.000,18.000) {-1};%
\draw[draw=crGrid,line width=0.45pt] (438.091,27.000) -- (438.091,181.000);%
\node[anchor=center,text=crInk,font=\fontsize{8}{9}\selectfont] at (438.091,18.000) {-0.5};%
\draw[draw=crGrid,line width=0.45pt] (467.182,27.000) -- (467.182,181.000);%
\node[anchor=center,text=crInk,font=\fontsize{8}{9}\selectfont] at (467.182,18.000) {0};%
\draw[draw=crInk,line width=0.5pt] (409.000,27.000) -- (473.000,27.000);%
\node[anchor=center,text=crInk,font=\fontsize{8}{9}\selectfont] at (450.000,6.000) {Change in (B - A)};%
\node[anchor=west,text=crInk,font=\fontsize{8}{9}\selectfont] at (1.000,169.000) {906 / wrong period};%
\draw[draw=crInk,line width=0.65pt] (220.850,174.000) -- (219.820,174.000);%
\draw[draw=crInk,fill=crInk,line width=.9pt] (220.850,175.100) circle (2.3pt);%
\draw[draw=crBlue,fill=white,line width=.9pt] (217.570,170.650) rectangle (222.070,175.150);%
\node[anchor=center,text=crInk,font=\fontsize{8}{9}\selectfont] at (174.500,165.000) {0.950 / 0.940};%
\draw[draw=crInk,line width=0.65pt] (356.850,174.000) -- (270.845,174.000);%
\draw[draw=crInk,fill=crInk,line width=.9pt] (356.850,175.100) circle (2.3pt);%
\draw[draw=crBlue,fill=white,line width=.9pt] (268.595,170.650) rectangle (273.095,175.150);%
\node[anchor=center,text=crInk,font=\fontsize{8}{9}\selectfont] at (310.500,165.000) {0.950 / 0.115};%
\draw[draw=crInk,line width=0.8pt] (419.182,172.000) -- (467.182,172.000);%
\draw[draw=crInk,fill=crInk,line width=.9pt] (419.182,172.000) circle (2.3pt);%
\node[anchor=east,text=crInk,font=\fontsize{8}{9}\selectfont] at (503.000,172.000) {-0.825};%
\node[anchor=west,text=crInk,font=\fontsize{8}{9}\selectfont] at (1.000,148.000) {930 / wrong period};%
\draw[draw=crInk,line width=0.65pt] (221.365,153.000) -- (220.335,153.000);%
\draw[draw=crInk,fill=crInk,line width=.9pt] (221.365,154.100) circle (2.3pt);%
\draw[draw=crBlue,fill=white,line width=.9pt] (218.085,149.650) rectangle (222.585,154.150);%
\node[anchor=center,text=crInk,font=\fontsize{8}{9}\selectfont] at (174.500,144.000) {0.955 / 0.945};%
\draw[draw=crInk,line width=0.65pt] (358.910,153.000) -- (280.630,153.000);%
\draw[draw=crInk,fill=crInk,line width=.9pt] (358.910,154.100) circle (2.3pt);%
\draw[draw=crBlue,fill=white,line width=.9pt] (278.380,149.650) rectangle (282.880,154.150);%
\node[anchor=center,text=crInk,font=\fontsize{8}{9}\selectfont] at (310.500,144.000) {0.970 / 0.210};%
\draw[draw=crInk,line width=0.8pt] (423.545,151.000) -- (467.182,151.000);%
\draw[draw=crInk,fill=crInk,line width=.9pt] (423.545,151.000) circle (2.3pt);%
\node[anchor=east,text=crInk,font=\fontsize{8}{9}\selectfont] at (503.000,151.000) {-0.750};%
\node[anchor=west,text=crInk,font=\fontsize{8}{9}\selectfont] at (1.000,127.000) {746 / wrong period};%
\draw[draw=crInk,line width=0.65pt] (215.700,132.000) -- (216.215,132.000);%
\draw[draw=crInk,fill=crInk,line width=.9pt] (215.700,133.100) circle (2.3pt);%
\draw[draw=crBlue,fill=white,line width=.9pt] (213.965,128.650) rectangle (218.465,133.150);%
\node[anchor=center,text=crInk,font=\fontsize{8}{9}\selectfont] at (174.500,123.000) {0.900 / 0.905};%
\draw[draw=crInk,line width=0.65pt] (357.880,132.000) -- (294.020,132.000);%
\draw[draw=crInk,fill=crInk,line width=.9pt] (357.880,133.100) circle (2.3pt);%
\draw[draw=crBlue,fill=white,line width=.9pt] (291.770,128.650) rectangle (296.270,133.150);%
\node[anchor=center,text=crInk,font=\fontsize{8}{9}\selectfont] at (310.500,123.000) {0.960 / 0.340};%
\draw[draw=crInk,line width=0.8pt] (430.818,130.000) -- (467.182,130.000);%
\draw[draw=crInk,fill=crInk,line width=.9pt] (430.818,130.000) circle (2.3pt);%
\node[anchor=east,text=crInk,font=\fontsize{8}{9}\selectfont] at (503.000,130.000) {-0.625};%
\node[anchor=west,text=crInk,font=\fontsize{8}{9}\selectfont] at (1.000,106.000) {930 / wrong status};%
\draw[draw=crInk,line width=0.65pt] (220.850,111.000) -- (220.335,111.000);%
\draw[draw=crInk,fill=crInk,line width=.9pt] (220.850,112.100) circle (2.3pt);%
\draw[draw=crBlue,fill=white,line width=.9pt] (218.085,107.650) rectangle (222.585,112.150);%
\node[anchor=center,text=crInk,font=\fontsize{8}{9}\selectfont] at (174.500,102.000) {0.950 / 0.945};%
\draw[draw=crInk,line width=0.65pt] (358.395,111.000) -- (322.860,111.000);%
\draw[draw=crInk,fill=crInk,line width=.9pt] (358.395,112.100) circle (2.3pt);%
\draw[draw=crBlue,fill=white,line width=.9pt] (320.610,107.650) rectangle (325.110,112.150);%
\node[anchor=center,text=crInk,font=\fontsize{8}{9}\selectfont] at (310.500,102.000) {0.965 / 0.620};%
\draw[draw=crInk,line width=0.8pt] (447.400,109.000) -- (467.182,109.000);%
\draw[draw=crInk,fill=crInk,line width=.9pt] (447.400,109.000) circle (2.3pt);%
\node[anchor=east,text=crInk,font=\fontsize{8}{9}\selectfont] at (503.000,109.000) {-0.340};%
\node[anchor=west,text=crInk,font=\fontsize{8}{9}\selectfont] at (1.000,85.000) {535 / equivalent};%
\draw[draw=crInk,line width=0.65pt] (210.550,90.000) -- (210.550,90.000);%
\draw[draw=crInk,fill=crInk,line width=.9pt] (210.550,91.100) circle (2.3pt);%
\draw[draw=crBlue,fill=white,line width=.9pt] (208.300,86.650) rectangle (212.800,91.150);%
\node[anchor=center,text=crInk,font=\fontsize{8}{9}\selectfont] at (174.500,81.000) {0.850 / 0.850};%
\draw[draw=crInk,line width=0.65pt] (344.490,90.000) -- (339.855,90.000);%
\draw[draw=crInk,fill=crInk,line width=.9pt] (344.490,91.100) circle (2.3pt);%
\draw[draw=crBlue,fill=white,line width=.9pt] (337.605,86.650) rectangle (342.105,91.150);%
\node[anchor=center,text=crInk,font=\fontsize{8}{9}\selectfont] at (310.500,81.000) {0.830 / 0.785};%
\draw[draw=crInk,line width=0.8pt] (464.564,88.000) -- (467.182,88.000);%
\draw[draw=crInk,fill=crInk,line width=.9pt] (464.564,88.000) circle (2.3pt);%
\node[anchor=east,text=crInk,font=\fontsize{8}{9}\selectfont] at (503.000,88.000) {-0.045};%
\node[anchor=west,text=crInk,font=\fontsize{8}{9}\selectfont] at (1.000,64.000) {583 / equivalent};%
\draw[draw=crInk,line width=0.65pt] (204.370,69.000) -- (205.915,69.000);%
\draw[draw=crInk,fill=crInk,line width=.9pt] (204.370,70.100) circle (2.3pt);%
\draw[draw=crBlue,fill=white,line width=.9pt] (203.665,65.650) rectangle (208.165,70.150);%
\node[anchor=center,text=crInk,font=\fontsize{8}{9}\selectfont] at (174.500,60.000) {0.790 / 0.805};%
\draw[draw=crInk,line width=0.65pt] (335.735,69.000) -- (339.855,69.000);%
\draw[draw=crInk,fill=crInk,line width=.9pt] (335.735,70.100) circle (2.3pt);%
\draw[draw=crBlue,fill=white,line width=.9pt] (337.605,65.650) rectangle (342.105,70.150);%
\node[anchor=center,text=crInk,font=\fontsize{8}{9}\selectfont] at (310.500,60.000) {0.745 / 0.785};%
\draw[draw=crInk,line width=0.8pt] (468.636,67.000) -- (467.182,67.000);%
\draw[draw=crInk,fill=crInk,line width=.9pt] (468.636,67.000) circle (2.3pt);%
\node[anchor=east,text=crInk,font=\fontsize{8}{9}\selectfont] at (503.000,67.000) {+0.025};%
\node[anchor=west,text=crInk,font=\fontsize{8}{9}\selectfont] at (1.000,43.000) {785 / unresolved boundary};%
\draw[draw=crInk,line width=0.65pt] (183.770,48.000) -- (182.740,48.000);%
\draw[draw=crInk,fill=crInk,line width=.9pt] (183.770,49.100) circle (2.3pt);%
\draw[draw=crBlue,fill=white,line width=.9pt] (180.490,44.650) rectangle (184.990,49.150);%
\node[anchor=center,text=crInk,font=\fontsize{8}{9}\selectfont] at (174.500,39.000) {0.590 / 0.580};%
\draw[draw=crInk,line width=0.65pt] (328.010,48.000) -- (301.745,48.000);%
\draw[draw=crInk,fill=crInk,line width=.9pt] (328.010,49.100) circle (2.3pt);%
\draw[draw=crBlue,fill=white,line width=.9pt] (299.495,44.650) rectangle (303.995,49.150);%
\node[anchor=center,text=crInk,font=\fontsize{8}{9}\selectfont] at (310.500,39.000) {0.670 / 0.415};%
\draw[draw=crInk,line width=0.8pt] (452.927,46.000) -- (467.182,46.000);%
\draw[draw=crInk,fill=crInk,line width=.9pt] (452.927,46.000) circle (2.3pt);%
\node[anchor=east,text=crInk,font=\fontsize{8}{9}\selectfont] at (503.000,46.000) {-0.245};%
\end{tikzpicture}%
\endgroup%

\caption{Explicit-column intervention: all two-call A/B means and interactions. Interaction is explicit (B minus A) minus original (B minus A). Negative values indicate greater separation; wrong roles call for separation, whereas equivalent facts call for support in both arms. Boundary labels remain unresolved. Means are descriptive; question 930 shares a technically repeated A input.}\label{fig:2}
\end{figure*}

Fig.~\ref{fig:2} shows every cell mean. The interaction is the enriched B-minus-A difference minus the original B-minus-A difference. A negative interaction indicates greater separation between the original and swapped citations; whether that separation is desirable depends on the semantic stratum.

For the four selected wrong-role contrasts, all eight B calls are accepted under the original rendering and withheld under the explicit rendering at 0.8. All corresponding A calls remain accepted in both conditions. Enriched A means range from 0.950 to 0.970, whereas enriched B means range from 0.115 to 0.620. The largest observed identical-input range is 0.05 in this repeat check. These matched observations support a representation effect on the tested period and status errors.

The equivalent controls reveal the cost to valid evidence. Both enriched B means are 0.785; one of their four individual calls is accepted at the inclusive cutoff. Question 583 also has an original A mean below 0.8, and enrichment lowers its A mean further, producing a positive interaction. Explicit labels thus change support for valid as well as wrong citations. The accounting/parser case retains its unresolved label. Coverage is limited to the seven column-only contrasts; the earlier row-changing opening/closing-balance controls were not rerun.

On these selected cases, explicit column labels increase sensitivity to period and status mismatches while lowering support for some equivalent facts. This representation-sensitive tradeoff motivates the fresh-source test that follows.

\subsection{Fresh source pages with a returned non-author review}\label{sec:3.4}

To test beyond previously observed failures, we construct a separate bank from the pinned FinQA train split. After excluding 178 pages used by earlier local experiments, a structural screen yields 116 eligible questions on 91 pages: both operands must be leading signed numbers in table cells, with another same-value cell for at least one operand. A fixed hash ranking selects 36 questions on 36 pages from 35 reports, one question per page. A source-only rule chooses the original citations and the first alternative same-number cell in row-major order. The follow-up uses program-derived constructed records to test controlled citation changes. These pages are new to our local experiments; natural model-error frequencies and training-data overlap are outside this sampling design.

Before inference we save the inputs, sample and original/explicit-column schedule. All 288 evaluation calls (two arms, two renderings, two repeats per question) and four format checks succeed. The prompt, model, column-field intervention and cutoffs are unchanged. One non-author reviewer returns labels for all 36 contrasts, declaring unaided reading of the supplied transcripts and no access to model outputs, author labels or the blinded X/Y mapping. Review took place after collection under the reviewer's declared blindness. The reviewer assesses support for each complete citation set, including its unchanged operand. The labels reflect one reviewer per case; inter-human agreement remains unmeasured.

The returned labels identify 15 contrasts with A supported and B unsupported, 19 with both supported, and two with neither supported. Eight contrast labels differ from the corrected assistant judgments (64 of 72 arm labels agree); all disagreements and original rationales remain saved. Fig.~\ref{fig:3} uses the returned labels as specified by the follow-up protocol; Tables 1--2 and Figures 1--2 retain their historical labels.

\begin{figure*}[!t]
\centering
\begingroup%
\definecolor{crInk}{HTML}{303942}%
\definecolor{crBlue}{HTML}{0072B2}%
\definecolor{crOrange}{HTML}{D55E00}%
\definecolor{crLight}{HTML}{F1F4F6}%
\definecolor{crGrid}{HTML}{DAE0E4}%
\begin{tikzpicture}[x=1pt,y=1pt,every node/.style={font=\fontsize{8}{9}\selectfont,text=crInk,inner sep=0pt,outer sep=0pt}]%
\path[use as bounding box] (0,0) rectangle (504,265);%
\node[anchor=west,text=crInk,font=\fontsize{8}{9}\selectfont] at (0.000,258.000) {(a) Acceptance outcomes at the inclusive 0.8 cutoff};%
\draw[draw=crInk,fill=crInk,line width=.9pt] (269.000,258.000) circle (2.3pt);%
\node[anchor=west,text=crInk,font=\fontsize{8}{9}\selectfont] at (275.000,258.000) {Original rendering};%
\draw[draw=crOrange,fill=white,line width=.9pt] (363.750,255.750) rectangle (368.250,260.250);%
\node[anchor=west,text=crInk,font=\fontsize{8}{9}\selectfont] at (373.000,258.000) {Explicit column};%
\node[anchor=center,text=crInk,font=\fontsize{8}{9}\selectfont] at (441.000,241.000) {Original};%
\node[anchor=center,text=crInk,font=\fontsize{8}{9}\selectfont] at (486.000,241.000) {Explicit};%
\draw[draw=crGrid,line width=0.45pt] (235.000,128.000) -- (235.000,236.000);%
\draw[draw=crInk,line width=0.5pt] (235.000,128.000) -- (235.000,126.000);%
\node[anchor=center,text=crInk,font=\fontsize{8}{9}\selectfont] at (235.000,119.000) {0};%
\draw[draw=crGrid,line width=0.45pt] (316.500,128.000) -- (316.500,236.000);%
\draw[draw=crInk,line width=0.5pt] (316.500,128.000) -- (316.500,126.000);%
\node[anchor=center,text=crInk,font=\fontsize{8}{9}\selectfont] at (316.500,119.000) {.5};%
\draw[draw=crGrid,line width=0.45pt] (398.000,128.000) -- (398.000,236.000);%
\draw[draw=crInk,line width=0.5pt] (398.000,128.000) -- (398.000,126.000);%
\node[anchor=center,text=crInk,font=\fontsize{8}{9}\selectfont] at (398.000,119.000) {1};%
\draw[draw=crInk,line width=0.5pt] (235.000,128.000) -- (398.000,128.000);%
\fill[crLight] (0.000,186.000) rectangle (504.000,232.000);%
\node[anchor=west,text=crInk,font=\fontsize{8}{9}\selectfont] at (0.000,214.000) {Only A supported};%
\node[anchor=west,text=crInk,font=\fontsize{8}{9}\selectfont] at (0.000,204.000) {(n = 15 pairs)};%
\draw[draw=crInk,line width=0.6pt] (101.000,186.000) -- (101.000,232.000);%
\draw[draw=crInk,line width=0.6pt] (101.000,186.000) -- (105.000,186.000);%
\draw[draw=crInk,line width=0.6pt] (101.000,232.000) -- (105.000,232.000);%
\node[anchor=west,text=crInk,font=\fontsize{8}{9}\selectfont] at (0.000,174.000) {Both supported};%
\node[anchor=west,text=crInk,font=\fontsize{8}{9}\selectfont] at (0.000,164.000) {(n = 19 pairs)};%
\draw[draw=crInk,line width=0.6pt] (101.000,154.000) -- (101.000,184.000);%
\draw[draw=crInk,line width=0.6pt] (101.000,154.000) -- (105.000,154.000);%
\draw[draw=crInk,line width=0.6pt] (101.000,184.000) -- (105.000,184.000);%
\fill[crLight] (0.000,138.000) rectangle (504.000,152.000);%
\node[anchor=west,text=crInk,font=\fontsize{8}{9}\selectfont] at (0.000,150.000) {Neither supported};%
\node[anchor=west,text=crInk,font=\fontsize{8}{9}\selectfont] at (0.000,140.000) {(n = 2 pairs)};%
\draw[draw=crInk,line width=0.6pt] (101.000,138.000) -- (101.000,152.000);%
\draw[draw=crInk,line width=0.6pt] (101.000,138.000) -- (105.000,138.000);%
\draw[draw=crInk,line width=0.6pt] (101.000,152.000) -- (105.000,152.000);%
\node[anchor=west,text=crInk,font=\fontsize{8}{9}\selectfont] at (112.000,225.000) {accept A, withhold B};%
\draw[draw=crInk,line width=0.75pt] (321.933,225.000) -- (332.800,225.000);%
\draw[draw=crInk,fill=crInk,line width=.9pt] (321.933,226.100) circle (2.3pt);%
\draw[draw=crOrange,fill=white,line width=.9pt] (330.550,221.650) rectangle (335.050,226.150);%
\node[anchor=center,text=crInk,font=\fontsize{8}{9}\selectfont] at (441.000,225.000) {8/15};%
\node[anchor=center,text=crOrange,font=\fontsize{8}{9}\selectfont] at (486.000,225.000) {9/15};%
\node[anchor=west,text=crInk,font=\fontsize{8}{9}\selectfont] at (112.000,209.000) {accept B};%
\draw[draw=crInk,line width=0.75pt] (256.733,209.000) -- (235.000,209.000);%
\draw[draw=crInk,fill=crInk,line width=.9pt] (256.733,210.100) circle (2.3pt);%
\draw[draw=crOrange,fill=white,line width=.9pt] (232.750,205.650) rectangle (237.250,210.150);%
\node[anchor=center,text=crInk,font=\fontsize{8}{9}\selectfont] at (441.000,209.000) {2/15};%
\node[anchor=center,text=crOrange,font=\fontsize{8}{9}\selectfont] at (486.000,209.000) {0/15};%
\node[anchor=west,text=crInk,font=\fontsize{8}{9}\selectfont] at (112.000,193.000) {withhold A};%
\draw[draw=crInk,line width=0.75pt] (289.333,193.000) -- (300.200,193.000);%
\draw[draw=crInk,fill=crInk,line width=.9pt] (289.333,194.100) circle (2.3pt);%
\draw[draw=crOrange,fill=white,line width=.9pt] (297.950,189.650) rectangle (302.450,194.150);%
\node[anchor=center,text=crInk,font=\fontsize{8}{9}\selectfont] at (441.000,193.000) {5/15};%
\node[anchor=center,text=crOrange,font=\fontsize{8}{9}\selectfont] at (486.000,193.000) {6/15};%
\node[anchor=west,text=crInk,font=\fontsize{8}{9}\selectfont] at (112.000,177.000) {withhold B};%
\draw[draw=crInk,line width=0.75pt] (329.368,177.000) -- (320.789,177.000);%
\draw[draw=crInk,fill=crInk,line width=.9pt] (329.368,178.100) circle (2.3pt);%
\draw[draw=crOrange,fill=white,line width=.9pt] (318.539,173.650) rectangle (323.039,178.150);%
\node[anchor=center,text=crInk,font=\fontsize{8}{9}\selectfont] at (441.000,177.000) {11/19};%
\node[anchor=center,text=crOrange,font=\fontsize{8}{9}\selectfont] at (486.000,177.000) {10/19};%
\node[anchor=west,text=crInk,font=\fontsize{8}{9}\selectfont] at (112.000,161.000) {accept both};%
\draw[draw=crInk,line width=0.75pt] (277.895,161.000) -- (286.474,161.000);%
\draw[draw=crInk,fill=crInk,line width=.9pt] (277.895,162.100) circle (2.3pt);%
\draw[draw=crOrange,fill=white,line width=.9pt] (284.224,157.650) rectangle (288.724,162.150);%
\node[anchor=center,text=crInk,font=\fontsize{8}{9}\selectfont] at (441.000,161.000) {5/19};%
\node[anchor=center,text=crOrange,font=\fontsize{8}{9}\selectfont] at (486.000,161.000) {6/19};%
\node[anchor=west,text=crInk,font=\fontsize{8}{9}\selectfont] at (112.000,145.000) {withhold both};%
\draw[draw=crInk,line width=0.75pt] (398.000,145.000) -- (398.000,145.000);%
\draw[draw=crInk,fill=crInk,line width=.9pt] (398.000,146.100) circle (2.3pt);%
\draw[draw=crOrange,fill=white,line width=.9pt] (395.750,141.650) rectangle (400.250,146.150);%
\node[anchor=center,text=crInk,font=\fontsize{8}{9}\selectfont] at (441.000,145.000) {2/2};%
\node[anchor=center,text=crOrange,font=\fontsize{8}{9}\selectfont] at (486.000,145.000) {2/2};%
\node[anchor=center,text=crInk,font=\fontsize{8}{9}\selectfont] at (316.500,110.000) {Fraction of the stated reviewer stratum};%
\node[anchor=west,text=crInk,font=\fontsize{8}{9}\selectfont] at (0.000,99.000) {(b) Accept A, withhold B; only A supported};%
\draw[draw=crGrid,line width=0.45pt] (35.000,39.000) -- (232.000,39.000);%
\node[anchor=east,text=crInk,font=\fontsize{8}{9}\selectfont] at (21.000,39.000) {0};%
\draw[draw=crGrid,line width=0.45pt] (35.000,63.000) -- (232.000,63.000);%
\node[anchor=east,text=crInk,font=\fontsize{8}{9}\selectfont] at (21.000,63.000) {0.5};%
\draw[draw=crGrid,line width=0.45pt] (35.000,87.000) -- (232.000,87.000);%
\node[anchor=east,text=crInk,font=\fontsize{8}{9}\selectfont] at (21.000,87.000) {1};%
\draw[draw=crInk,line width=0.5pt] (35.000,39.000) -- (35.000,37.000);%
\node[anchor=center,text=crInk,font=\fontsize{8}{9}\selectfont] at (35.000,18.000) {0.5};%
\draw[draw=crInk,line width=0.5pt] (166.333,39.000) -- (166.333,37.000);%
\node[anchor=center,text=crInk,font=\fontsize{8}{9}\selectfont] at (166.333,18.000) {0.8};%
\draw[draw=crInk,line width=0.5pt] (210.111,39.000) -- (210.111,37.000);%
\node[anchor=center,text=crInk,font=\fontsize{8}{9}\selectfont] at (210.111,18.000) {0.9};%
\draw[draw=crInk,line width=0.5pt] (232.000,39.000) -- (232.000,37.000);%
\node[anchor=center,text=crInk,font=\fontsize{8}{9}\selectfont] at (232.000,18.000) {0.95};%
\draw[draw=crInk,line width=0.9pt] (35.000,64.600) -- (166.333,64.600);%
\draw[draw=crInk,line width=0.9pt] (166.333,64.600) -- (210.111,58.200);%
\draw[draw=crInk,line width=0.9pt] (210.111,58.200) -- (232.000,45.400);%
\draw[draw=crInk,fill=crInk,line width=.9pt] (35.000,64.600) circle (2.3pt);%
\node[anchor=center,text=crInk,font=\fontsize{8}{9}\selectfont] at (35.000,56.600) {8/15};%
\draw[draw=crInk,fill=crInk,line width=.9pt] (166.333,64.600) circle (2.3pt);%
\node[anchor=center,text=crInk,font=\fontsize{8}{9}\selectfont] at (166.333,56.600) {8/15};%
\draw[draw=crInk,fill=crInk,line width=.9pt] (210.111,58.200) circle (2.3pt);%
\node[anchor=center,text=crInk,font=\fontsize{8}{9}\selectfont] at (210.111,50.200) {6/15};%
\draw[draw=crInk,fill=crInk,line width=.9pt] (232.000,45.400) circle (2.3pt);%
\node[anchor=center,text=crInk,font=\fontsize{8}{9}\selectfont] at (232.000,37.400) {2/15};%
\draw[draw=crOrange,line width=0.9pt,densely dashed] (35.000,67.800) -- (166.333,67.800);%
\draw[draw=crOrange,line width=0.9pt,densely dashed] (166.333,67.800) -- (210.111,61.400);%
\draw[draw=crOrange,line width=0.9pt,densely dashed] (210.111,61.400) -- (232.000,48.600);%
\draw[draw=crOrange,fill=white,line width=.9pt] (32.750,65.550) rectangle (37.250,70.050);%
\node[anchor=center,text=crOrange,font=\fontsize{8}{9}\selectfont] at (35.000,75.800) {9/15};%
\draw[draw=crOrange,fill=white,line width=.9pt] (164.083,65.550) rectangle (168.583,70.050);%
\node[anchor=center,text=crOrange,font=\fontsize{8}{9}\selectfont] at (166.333,75.800) {9/15};%
\draw[draw=crOrange,fill=white,line width=.9pt] (207.861,59.150) rectangle (212.361,63.650);%
\node[anchor=center,text=crOrange,font=\fontsize{8}{9}\selectfont] at (210.111,69.400) {7/15};%
\draw[draw=crOrange,fill=white,line width=.9pt] (229.750,46.350) rectangle (234.250,50.850);%
\node[anchor=center,text=crOrange,font=\fontsize{8}{9}\selectfont] at (232.000,56.600) {3/15};%
\node[anchor=center,text=crInk,font=\fontsize{8}{9}\selectfont] at (133.500,5.000) {Inclusive acceptance cutoff};%
\node[anchor=west,text=crInk,font=\fontsize{8}{9}\selectfont] at (262.000,99.000) {(c) Accept both; both supported};%
\draw[draw=crGrid,line width=0.45pt] (297.000,39.000) -- (494.000,39.000);%
\node[anchor=east,text=crInk,font=\fontsize{8}{9}\selectfont] at (283.000,39.000) {0};%
\draw[draw=crGrid,line width=0.45pt] (297.000,63.000) -- (494.000,63.000);%
\node[anchor=east,text=crInk,font=\fontsize{8}{9}\selectfont] at (283.000,63.000) {0.5};%
\draw[draw=crGrid,line width=0.45pt] (297.000,87.000) -- (494.000,87.000);%
\node[anchor=east,text=crInk,font=\fontsize{8}{9}\selectfont] at (283.000,87.000) {1};%
\draw[draw=crInk,line width=0.5pt] (297.000,39.000) -- (297.000,37.000);%
\node[anchor=center,text=crInk,font=\fontsize{8}{9}\selectfont] at (297.000,18.000) {0.5};%
\draw[draw=crInk,line width=0.5pt] (428.333,39.000) -- (428.333,37.000);%
\node[anchor=center,text=crInk,font=\fontsize{8}{9}\selectfont] at (428.333,18.000) {0.8};%
\draw[draw=crInk,line width=0.5pt] (472.111,39.000) -- (472.111,37.000);%
\node[anchor=center,text=crInk,font=\fontsize{8}{9}\selectfont] at (472.111,18.000) {0.9};%
\draw[draw=crInk,line width=0.5pt] (494.000,39.000) -- (494.000,37.000);%
\node[anchor=center,text=crInk,font=\fontsize{8}{9}\selectfont] at (494.000,18.000) {0.95};%
\draw[draw=crInk,line width=0.9pt] (297.000,71.842) -- (428.333,51.632);%
\draw[draw=crInk,line width=0.9pt] (428.333,51.632) -- (472.111,39.000);%
\draw[draw=crInk,line width=0.9pt] (472.111,39.000) -- (494.000,39.000);%
\draw[draw=crInk,fill=crInk,line width=.9pt] (297.000,71.842) circle (2.3pt);%
\node[anchor=center,text=crInk,font=\fontsize{8}{9}\selectfont] at (297.000,63.842) {13/19};%
\draw[draw=crInk,fill=crInk,line width=.9pt] (428.333,51.632) circle (2.3pt);%
\node[anchor=center,text=crInk,font=\fontsize{8}{9}\selectfont] at (428.333,43.632) {5/19};%
\draw[draw=crInk,fill=crInk,line width=.9pt] (472.111,39.000) circle (2.3pt);%
\node[anchor=center,text=crInk,font=\fontsize{8}{9}\selectfont] at (472.111,31.000) {0/19};%
\draw[draw=crInk,fill=crInk,line width=.9pt] (494.000,39.000) circle (2.3pt);%
\node[anchor=center,text=crInk,font=\fontsize{8}{9}\selectfont] at (494.000,31.000) {0/19};%
\draw[draw=crOrange,line width=0.9pt,densely dashed] (297.000,69.316) -- (428.333,54.158);%
\draw[draw=crOrange,line width=0.9pt,densely dashed] (428.333,54.158) -- (472.111,41.526);%
\draw[draw=crOrange,line width=0.9pt,densely dashed] (472.111,41.526) -- (494.000,39.000);%
\draw[draw=crOrange,fill=white,line width=.9pt] (294.750,67.066) rectangle (299.250,71.566);%
\node[anchor=center,text=crOrange,font=\fontsize{8}{9}\selectfont] at (297.000,77.316) {12/19};%
\draw[draw=crOrange,fill=white,line width=.9pt] (426.083,51.908) rectangle (430.583,56.408);%
\node[anchor=center,text=crOrange,font=\fontsize{8}{9}\selectfont] at (428.333,62.158) {6/19};%
\draw[draw=crOrange,fill=white,line width=.9pt] (469.861,39.276) rectangle (474.361,43.776);%
\node[anchor=center,text=crOrange,font=\fontsize{8}{9}\selectfont] at (472.111,49.526) {1/19};%
\draw[draw=crOrange,fill=white,line width=.9pt] (491.750,36.750) rectangle (496.250,41.250);%
\node[anchor=center,text=crOrange,font=\fontsize{8}{9}\selectfont] at (494.000,47.000) {0/19};%
\node[anchor=center,text=crInk,font=\fontsize{8}{9}\selectfont] at (395.500,5.000) {Inclusive acceptance cutoff};%
\end{tikzpicture}%
\endgroup%

\caption{Fresh-source outcomes under one reviewer's declared blinded labels. Decisions use two-call means and inclusive cutoffs. (a) All six outcomes at 0.8; overlapping measurements within disjoint strata of 15 A-only-supported, 19 both-supported and 2 neither-supported pairs. (b) Accept A, withhold B among A-only-supported pairs. (c) Accept both among both-supported pairs. Cutoffs were fixed before collection; lines guide the eye. A is rule-selected; B changes one same-number pointer. Fractions describe this constructed bank.}\label{fig:3}
\end{figure*}

The follow-up quantifies the tradeoff between wrong-role separation and supported-evidence coverage. Explicit columns remove the two unsupported-B acceptances, but supported-A withholding rises from five to six; paired separation increases by one contrast. Among equivalent controls, most pairs still have at least one arm withheld. With numerical agreement held fixed, these contrasts reveal both successful role separation and withholding of valid evidence.

Continuous scores qualify this policy-level interpretation. Among the 53 reviewer-supported arms, 25 are withheld at 0.8 under original rendering and 24 under explicit rendering; respectively 16 and 13 of those still have means at least 0.5. Means below 0.5 occur on 9 and 11 supported arms. In the 15 A-only-supported contrasts, mean A exceeds mean B in 11 versus 14 cases, but median A-minus-B decreases from 0.500 to 0.440. Thus ordering, separation magnitude and acceptance coverage are distinct readouts. The saved per-case ledger reports all 288 probabilities, two-call means and paired differences; Fig.~\ref{fig:3} reports the complete original cutoff set.

All six column-only contrasts remain visible: four reviewer-labeled wrong roles and two equivalents. Explicit headers preserve accepted A while withholding B on two wrong-role examples (444 and 4296); the other two wrong-role pairs have both arms withheld in both renderings. Equivalent case 1687 changes from both accepted to B withheld, while equivalent case 3921 remains accepted. These three threshold-changing examples illustrate a local benefit and its specificity cost within the full six-contrast subgroup. Case 444's issued/outstanding distinction also admits a contextual-equivalence reading. Post hoc exclusion of this case removes the net gain: paired separation is 8/14 in both renderings. Excluding all eight author/reviewer disagreements likewise gives 8/14 in both renderings, with both-equivalent acceptance at 3/12 versus 4/12. These sensitivity analyses retain the original labels.

All 36 review cases reproduce the question and source text in the verifier inputs, but the reviewer can inspect both arms and the whole packet. A source-only check finds one rationale, for case 1185, explicitly borrowing another case's year-column order. Excluding it gives paired separation 8/14 versus 9/14 and leaves the 19 equivalent pairs unchanged. A broader post hoc header-quality exclusion covers 1185, 2484 (duplicated year labels), and 3521 and 2829 (data rows reused as headers); separation is 6/12 versus 9/12 and both-equivalent acceptance is 5/18 versus 6/18. Case 4296 remains because its own prose supplies the year order. These source-quality sensitivities retain the existing review, with the full 36-case analysis remaining primary.

The four semantic concerns recorded before inference remain a sensitivity, even though the reviewer labels all four as equivalent. Excluding them leaves the negative readout unchanged and equivalent B withholding at 9/15 versus 8/15. Excluding headerless case 3521 changes negative paired separation to 7/14 versus 9/14; its original A mean 0.805 becomes 0.795, so its lost acceptance is a small cutoff crossing. The largest identical-input repeat range is 0.18. The case-level results expose sensitivity to role interpretation and near-threshold means; they describe this constructed bank rather than population rates.

\section{Verification and answer-release outcomes}\label{sec:4}

Cascade analysis emphasizes the conditional quality of both the primary model and its fallback \cite{r7}. The verifier experiments measure support probabilities on saved traces. Table~\ref{tab:3} evaluates answer release on the same question bank under the original consultation rule introduced in Section~\ref{sec:2}, retaining its prospectively specified endpoint and full denominator.

For this original task, Jev chooses among five numerical candidates: the rounded key, a one-sided neighbor and three arithmetic distractors assigned to neutral IDs by a fixed hash. The neighbor gap is the larger of one rounding unit and a rounded 5\% of the absolute key; actual relative gaps span 4.08\%--100\%. These transformed questions define a FinQA-derived selection task. Trace generation uses menu-free inputs.

The inherited policy consults the saved trace only when the maximum Choice probability is strictly below 0.8, otherwise retaining Jev's answer. Stated and executed fallback are compared using the same trace; a secondary policy requires literal/unit acceptance or abstains. This consultation rule concerns numerical selection and is distinct from the inclusive source-support gates in Table~\ref{tab:2}. The later Noul verifier and column-rendering intervention are not inserted into this original policy.

\begin{table*}[!t]
\centering
\caption{Prospectively specified financial release comparison.}
\label{tab:3}
\small
\setlength{\tabcolsep}{5pt}
\renewcommand{\arraystretch}{1.12}
\par\smallskip\noindent Each cell gives key matches / mismatches / unavailable among all 96 questions. The same primary response is used in both trace draws; the mean is computed within question. The gate is strictly below 0.8.\par\smallskip
\begin{tabularx}{\textwidth}{@{}>{\raggedright\arraybackslash}Xrrrr@{}}
\toprule
\textbf{Release policy} & \textbf{Trace draw 0} & \textbf{Trace draw 1} & \textbf{\shortstack[r]{Mean net match\\change (pp)}} & \textbf{\shortstack[r]{Consulted\\questions}} \\
\midrule
Jev alone & 95 / 1 / 0 & 95 / 1 / 0 & +0.00 & 0 \\
Gate \ensuremath{\rightarrow} stated number & 93 / 3 / 0 & 94 / 2 / 0 & -1.56 & 12 \\
Gate \ensuremath{\rightarrow} executed number & 93 / 3 / 0 & 94 / 2 / 0 & -1.56 & 12 \\
Gate \ensuremath{\rightarrow} execution + lexical/unit acceptance & 88 / 1 / 7 & 87 / 1 / 8 & -7.81 & 12 \\
\bottomrule
\end{tabularx}

\end{table*}

Jev matches 95 of 96 keys. All 12 consulted questions already have correct primary answers; the sole primary error has probability exactly 0.8 and is outside the strict gate. Thus the observed gate has no correction opportunity in this bank. Execution corrects 13 stated outputs on seven questions across the complete trace bank, but none of those corrections activates the gate. Two execution regressions share a separate question involving output-unit scaling.

Executed fallback yields 93 and 94 key matches across the two trace draws. The prospectively specified mean within-question difference is -1.5625 percentage points, with a descriptive 95\% question-bootstrap interval of [-4.1667, 0] from 20,000 resamples. The interval includes zero, so a zero population effect remains compatible with this descriptive estimate. A post hoc sensitivity accepting both readings of a temporal ambiguity changes the point difference to -1.04 points without replacing the original scoring. The original fixed-gate comparison provides a measured workflow boundary for the component findings: no release benefit was observed on this bank, and the later source verifier was evaluated separately.

\section{Implications and relation to prior work}\label{sec:5}

\subsection{Incremental information and failure conditions}\label{sec:5.1}

The natural-record comparison identifies the value of a strong numerical baseline: signed-number matching explains most recovery over exact quotation checks, while probabilistic verification offers different coverage and joint-pass tradeoffs across cutoffs. Same-number swaps then expose citation distinctions that numerical agreement alone cannot resolve. Jev responds to some wrong roles but misses others, and the acceptance policy also withholds some equivalent citations. The incremental-information question therefore requires both unsupported-role detection and supported-evidence retention.

The row/column response difference motivates a controlled test of representation sensitivity. Making column identity explicit changes the tested period/status decisions, while equivalent controls distinguish justified rejection from withholding of valid evidence. The fresh-source follow-up retains this tradeoff under a non-author reviewer and reveals withholding of supported original citations at the chosen cutoff. Together, these comparisons identify sensitivity to the supplied representation and its acceptance-coverage consequences.

For financial research assistants, verification must check the link between an LLM-generated operand and documentary evidence. Tool execution checks the proposed arithmetic; citation verification additionally assesses the operand's financial role. A verifier evaluation should test both rejection of unsupported roles and retention of equivalent evidence, such as carried-forward balances. Our matched inputs hold arithmetic fixed and evaluate emitted records, making cited-role support a distinct target alongside numerical correctness and answer release. Section~\ref{sec:4} retains the original fixed-gate release comparison as the workflow-level endpoint.

\subsection{Closest precedents and scope of contribution}\label{sec:5.2}

JEV-as-a-Judge is a direct precedent for probabilistic judging and escalation \cite{r13}. Our experiments examine what a judge distinguishes in cited financial evidence. BudgetVerify studies the allocation of verification effort in financial QA \cite{r14}. Equal-number and equivalent-fact controls add a complementary evaluation axis: whether a verifier distinguishes unsupported roles from valid alternative citations.

Hume's practitioner report probes Jev's contextual behavior \cite{r3}. This provides a black-box precedent for examining sensitivity to the supplied context. Sun et al. study option-name binding in typed decision models \cite{r4}. Here the controlled variation concerns cited financial roles at fixed numerical values.

VeriFin combines financial evidence grounding with symbolic verification \cite{r8}. Our target is the probability assigned to support at a specified citation. FinCARDS aligns financial entities, metrics, periods and numerical spans for evidence reranking \cite{r9}. Checking the role of an already cited operand complements these grounding and reranking stages.

FinRank likewise studies evidence-grounded financial question answering and retrieval over SEC filings \cite{r10}. The present study holds the available source fixed and varies a citation or its rendering. This controlled comparison connects numerical baselines, financial-role contrasts and equivalent-fact controls within the evaluation of a probabilistic verifier.

\section{Limitations and reproducibility}\label{sec:6}

The natural bank is selected for single-operation calculations and spans overlapping company/year reports. The primary selector's 95/96 key matches leave little room for the original gate to improve accuracy. Source PDFs, training contamination and historical information cutoffs are outside the evaluation's scope. Historical labels come from unblinded assistant review. The fresh-source extension adds one returned non-author review, with blindness and method declared in its form; this is a single-reviewer assessment without inter-human reliability estimates or consensus adjudication. Temporal interpretation, malformed year headers, price-role completeness and source transcription defects require annotation judgment. One returned rationale explicitly infers missing year-column order from another supplied case, a context difference from the verifier's per-question input; Section~\ref{sec:3.4} reports its exclusion and a separate header-quality sensitivity. These analyses show dependence on label and source-quality choices; semantic label validity remains a separate requirement.

The historical extensions are exploratory: the source verifier was specified after inspecting natural records, the relaxed deterministic baseline was added after inspecting verifier results, and the column intervention revisits observed failures in constructed citation swaps. Inputs and assistant labels were saved before each collection within this exploratory sequence. The 36-question follow-up avoids selection by verifier outcomes, while its structurally filtered, program-derived population limits generalization. Four negative rendering contrasts span only three source questions, and question 930 shares an A input. The two equivalent B means in the selected column-rendering study sit close to the cutoff, making these decisions sensitive to small probability changes. Counts describe the selected cases; failure prevalence in financial reports remains unmeasured.

The experiments evaluate one returned Jev identity and one generative alias. Technical repeats and two trace draws characterize observed serving variability; future serving behavior requires separate assessment. Source probabilities target cited-role support against the stated annotation source. Trading performance, live-agent improvement and the model's internal reasoning are outside the evaluated endpoints.

Saved records retain source revisions, selection decisions, prompts, serialized inputs, model/provider identities, exact probabilities, technical-repeat groups, costs and source-review reasons. Original endpoints and historical experiments retain their own records. Recorded costs, rounded to five decimal places, are USD 0.14781 for the 288-call natural-bank collection, USD 0.01669 for the 196-call source-verifier extension, USD 0.00488 for 62 swap-study calls including format checks, USD 0.00481 for the 60-call rendering experiment, and USD 0.02417 for the 292-call fresh-source follow-up. These charges document observed provider billing; the experiments use different tasks and call budgets. Offline numerical and input-invariant checks support numerical reproducibility; semantic assessment relies on the retained source review. A portable inspection packet retains exact instructions, source contexts, paired inputs, individual responses, label reasons and the complete natural-record disagreement table. The blinded packet and exact returned form preserve the follow-up review and declaration. A follow-up ledger retains all 288 probabilities, original-cutoff summaries and post hoc label/header sensitivities. The inspection artifacts are retained privately; public access remains pending.

\section{Conclusion}\label{sec:7}

For LLM-generated financial calculations, probabilistic evidence verification should be assessed by what it adds beyond numerical matching and by the citation changes under which it fails. On the retained natural records, a signed-number-at-pointer rule explains most recovery over exact quotation checks. Matched-number swaps reveal a separate requirement: distinguishing unsupported financial roles from alternative citations to equivalent facts. At the stated cutoff, Jev accepts some of the former and withholds some of the latter despite strong aggregate discrimination on natural records.

The fresh-source follow-up, with a separate reviewer, extends this evaluation to new pages and highlights the tradeoff between detecting unsupported roles and retaining valid citations. The contribution is a controlled evaluation combining numerical baselines, same-number swaps and equivalent-fact controls, with column rendering used to probe representation sensitivity. For LLM-based financial assistants, the evaluation separates numerical correctness, cited-role support and answer-release quality.

\section*{ACKNOWLEDGMENT}

CodeX(OpenAI) was used for language editing, experiment running, and assistance in script development. The authors reviewed the manuscripts, verified the reported results and citations, and take responsibility for the final text.

\clearpage
\twocolumn[\section*{References}]
\begingroup
\renewcommand{\section}[2]{}
\IEEEtriggeratref{8}

\endgroup
\end{document}